**Addressing the Gaps in Early Dementia Detection:**

**A Path Towards Enhanced Diagnostic Models through Machine Learning.**

Author:

Juan A. **Berrios Moya**

juan.berriosmoya@live.vu.edu.au

Running Title:

Early Dementia Detection with Machine Learning.

**Abstract**

The rapid global aging trend has led to an increase in dementia cases, including Alzheimer's disease, underscoring the urgent need for early and accurate diagnostic methods. Traditional diagnostic techniques, such as cognitive tests, neuroimaging, and biomarker analysis, face significant limitations in sensitivity, accessibility, and cost, particularly in the early stages. This study explores the potential of machine learning (ML) as a transformative approach to enhance early dementia detection by leveraging ML models to analyze and integrate complex multimodal datasets, including cognitive assessments, neuroimaging, and genetic information. A comprehensive review of existing literature was conducted to evaluate various ML models, including supervised learning, deep learning, and advanced techniques such as ensemble learning and transformer models, assessing their accuracy, interpretability, and potential for clinical integration. The findings indicate that while ML models show significant promise in improving diagnostic precision and enabling earlier interventions, challenges remain in their generalizability, interpretability, and ethical deployment. This research concludes by outlining future directions aimed at enhancing the clinical utility of ML models in dementia detection, emphasizing interdisciplinary collaboration and ethically sound frameworks to improve early detection and intervention strategies for Alzheimer's disease and other forms of dementia.

**Introduction**

The global demographic shift toward an aging population represents one of the most profound challenges to public health in the 21st century. The United Nations Department of Economic and Social Affairs projects that the population aged 65 and over will reach 1.5 billion by 2050[1]. This unprecedented growth is paralleled by a significant rise in age-related diseases, with dementia emerging as one of the most critical concerns. The World Health Organization estimates that around 55 million people worldwide currently live with dementia, a number expected to triple by 2050[2], emphasizing the urgent need for effective early detection and intervention strategies.

Dementia, encompassing disorders such as Alzheimer's disease, is an irreversible neurodegenerative condition that progressively deteriorates cognitive and functional abilities, primarily affecting older adults[3]. Despite the absence of a cure, early detection of dementia offers substantial benefits. Interventions initiated at the earliest stages can delay cognitive decline, optimize long-term care planning, and significantly enhance the quality of life for both patients and caregivers[4]. However, the current landscape of dementia diagnosis faces significant obstacles, particularly in accurately identifying the disease in its nascent stages.

Traditional diagnostic methods, including cognitive assessments, neuroimaging, and biomarkers, while valuable, are fraught with limitations. Cognitive tests like the Mini-Mental State Examination (MMSE) and the Montreal Cognitive Assessment (MoCA), though widely used, often lack the sensitivity needed to detect early-stage cognitive impairments, especially in diverse populations with varying educational backgrounds[5] Neuroimaging techniques such as magnetic resonance imaging (MRI) and positron emission tomography (PET) provide detailed insights into brain structure and function but are hindered by high costs, limited accessibility, and the need for specialized interpretation[6] . Moreover, while biomarkers and genetic testing offer promising avenues for early detection, their application is limited by the invasiveness of procedures and the need for further validation and standardization in clinical practice[7].

In this context, machine learning (ML) presents a transformative opportunity to address these diagnostic challenges. ML techniques, with their ability to process and analyze large, complex datasets, have the potential to revolutionize dementia detection by enhancing the accuracy and accessibility of diagnostic models. This thesis explores the integration of ML into early dementia detection, focusing on how these technologies can leverage multimodal data—spanning cognitive tests, neuroimaging, and genetic information—to improve diagnostic precision and enable earlier interventions.

Through a comprehensive review of existing literature, this thesis evaluates a wide range of ML models, including supervised learning, unsupervised learning, deep learning approaches, and recurrent neural networks. Each of these models is examined in the context of their application to dementia diagnosis, with particular attention given to their strengths, limitations, and potential for clinical integration. The study also delves into advanced ML techniques, such as ensemble learning models, transformer models, and generative adversarial networks (GANs), assessing their capacity to handle the complexities inherent in dementia diagnosis.

Despite the promising advancements, this thesis identifies several critical gaps in the current research landscape. These include the need for improved interpretability of ML models, which often operate as "black boxes," making it difficult for clinicians to trust and utilize these tools in practice. Furthermore, the thesis highlights the importance of integrating temporal data and longitudinal studies to better predict the progression of cognitive decline. Ethical considerations are also a central focus, with an emphasis on the need for frameworks that ensure data privacy, mitigate algorithmic bias, and promote fairness in the deployment of ML models across diverse populations.

The findings and discussions presented in this thesis aim to provide a roadmap for future research and development in the field of dementia detection. By advancing the capabilities of ML models and addressing the ethical and practical challenges associated with their use, this work seeks to contribute to the global effort to improve early detection and intervention strategies for dementia. Ultimately, the goal is to foster the development of diagnostic tools that are not only more accurate and accessible but also ethically responsible, thereby enhancing the overall quality of dementia care and patient outcomes.

**Literature review**

*Overview of Dementia and Early Detection*

Dementia is a broad term encompassing a range of neurodegenerative disorders characterized by a progressive decline in cognitive function, severely impacting memory, thinking, orientation, comprehension, calculation, learning capacity, language, and judgment. Alzheimer's disease is the most common form of dementia, accounting for 60-80% of cases[8]. The onset of dementia is typically gradual, beginning with mild cognitive impairment (MCI) that can progress to more severe stages, significantly affecting daily life and requiring comprehensive care.

**Stages of Dementia**

Dementia progresses through several stages, each characterized by increasingly severe cognitive decline. In the early stage, individuals may experience subtle memory lapses, often dismissed as normal aging. As the disease progresses to the middle stage, cognitive impairments become more pronounced, affecting daily activities and personal care. In the late stage, individuals typically require full-time assistance as they lose the ability to communicate, recognize loved ones, and perform basic functions[2]. Understanding these stages is critical for early detection and intervention, which can slow the progression and improve quality of life[4].

**Importance of Early Detection**

Early detection of dementia is crucial as it allows for timely intervention, which can significantly alter the disease's trajectory. While there is currently no cure for dementia, early diagnosis can enable patients to benefit from existing treatments that may temporarily slow cognitive decline, manage symptoms more effectively, and improve overall well-being[9]. Additionally, early detection allows patients and their families to plan for the future, access support services, and participate in clinical trials for new treatments[10].

Studies have shown that the early stages of dementia, particularly MCI, present a window of opportunity for interventions that could delay the onset of more severe symptoms. For instance, interventions such as cognitive stimulation therapy, lifestyle changes, and pharmacological treatments may be more effective when implemented early[11].Moreover, early diagnosis can help mitigate the economic burden on healthcare systems by delaying the need for more intensive care services[4].

**Cognitive Tests for Early Detection**

Several cognitive tests are routinely used to screen for early signs of dementia, with the Mini-Mental State Examination (MMSE) and the Montreal Cognitive Assessment (MoCA) being among the most widely employed.

**Mini-Mental State Examination (MMSE):** Developed by Folstein[12], the MMSE is a 30-point questionnaire that assesses various cognitive functions, including arithmetic, memory, and orientation. The MMSE is widely used due to its simplicity and effectiveness in detecting cognitive impairment. A score below 24 out of 30 typically indicates cognitive impairment, although this can vary based on age, education, and cultural background.

**Montreal Cognitive Assessment (MoCA):** The MoCA was developed to provide a more sensitive tool for detecting MCI, which often precedes dementia. It includes tasks that assess short-term memory, visuospatial abilities, executive function, attention, language, and orientation. The MoCA has been shown to have higher sensitivity than the MMSE in detecting early cognitive changes, particularly in highly educated individuals[5].

Both tests are widely used in clinical settings and research to identify individuals at risk of developing dementia. However, they have limitations, particularly in terms of their ability to detect very early stages of cognitive decline and their reliance on cut-off scores that may not account for cultural and educational differences[13].

**Global Impact of Dementia**

Dementia's impact on global health is profound, with significant implications for individuals, families, and healthcare systems. The World Health Organization has recognized dementia as a public health priority, urging member states to develop comprehensive national strategies to address the growing challenge.

As of 2019, dementia is one of the leading causes of disability and dependency among older adults, with significant social and economic costs. The total global cost of dementia was estimated at US$1 trillion in 2018, a figure expected to double by 2030 due to the aging population[1]. These costs include direct medical care, social care, and informal care provided by families, which often results in financial strain and reduced quality of life for caregivers.

Given the growing prevalence and economic burden of dementia, there is an urgent need for more effective early detection methods. Early intervention could significantly reduce the long-term costs associated with dementia care and improve outcomes for millions of individuals worldwide[14].

*Current Diagnostic Methods and Limitations*

The diagnosis of dementia is a complex process that requires the integration of various diagnostic tools and clinical judgment. The primary goal is to identify cognitive impairment early in the disease's progression to implement appropriate interventions.

However, the current diagnostic methods, while useful, have several limitations that hinder their effectiveness in routine clinical practice and broad public health screening.

**Cognitive Tests: Strengths and Weaknesses**

Cognitive tests are the cornerstone of dementia diagnosis. Among the most widely used are the Mini-Mental State Examination (MMSE) and the Montreal Cognitive Assessment (MoCA). These tests are designed to assess various cognitive domains, including memory, attention, language, visuospatial skills, and executive function.

**Mini-Mental State Examination (MMSE):** The MMSE has several limitations. It is less sensitive in detecting early cognitive impairment, especially in highly educated individuals, and it is also influenced by age, education, and cultural background, which can result in false negatives or positives[15]. Additionally, it lacks the sensitivity to differentiate between different types of dementia, such as Alzheimer's disease and vascular dementia[16].

**Montreal Cognitive Assessment (MoCA):** The MoCA was developed to address some of the limitations of the MMSE, particularly its insensitivity to MCI. The MoCA includes tasks that are more challenging and cover a broader range of cognitive domains, such as executive function and abstraction, which are often affected in the early stages of dementia[5]. Research has shown that the MoCA is more sensitive than the MMSE in detecting early cognitive changes, particularly in individuals with higher education levels[13]. However, similar to the MMSE, the MoCA's scores can be influenced by education and cultural factors, and its administration time is slightly longer, which can be a drawback in busy clinical settings[17].

**Clock Drawing Test (CDT):** The CDT is a simple and quick test often used in conjunction with other cognitive assessments. It evaluates a person's visuospatial ability and executive function by asking them to draw a clock showing a specific time. While the CDT is a valuable tool for assessing cognitive impairment, it has limitations in terms of sensitivity, especially in detecting mild cases of dementia[18]. The interpretation of the CDT can also be subjective, which may lead to variability in scoring[19].

**Test of Fluency (Verbal Fluency Tests):** These tests assess executive function by asking the patient to generate as many words as possible from a specific category (e.g., animals) or starting with a specific letter within a limited time. While useful in identifying cognitive deficits, particularly in the frontal lobe, they are less effective in detecting early stages of dementia and are often used as part of a broader assessment battery[20].

Despite the widespread use of these cognitive tests, their limitations underscore the need for more comprehensive tools and methods that can provide a more accurate diagnosis, especially in the early stages of dementia.

**Neuroimaging Techniques: Advantages and Challenges**

Neuroimaging techniques such as Magnetic Resonance Imaging (MRI) and Positron Emission Tomography (PET) have become critical components in the diagnosis of dementia. These techniques provide detailed images of the brain's structure and function, allowing for the detection of abnormalities associated with different types of dementia.

**Magnetic Resonance Imaging (MRI):** MRI is widely used to identify structural changes in the brain that are characteristic of neurodegenerative diseases. It can detect brain atrophy, particularly in regions like the hippocampus, which is often affected in Alzheimer's disease. MRI is non-invasive and does not involve radiation, making it a preferred method for repeated assessments. However, its limitations include high costs, limited availability in certain regions, and the fact that it requires significant expertise to interpret the results accurately[21].

**Positron Emission Tomography (PET):** PET imaging, especially using tracers like fluorodeoxyglucose (FDG) and amyloid PET, provides functional information about the brain. FDG-PET can identify areas of reduced glucose metabolism, which correlates with neuronal dysfunction, while amyloid PET can detect the accumulation of amyloid plaques, a hallmark of Alzheimer's disease[22]-[23]. While PET imaging is highly informative, it is expensive, involves exposure to radioactive substances, and is not widely available, which limits its routine use[24].

**Computed Tomography (CT):** CT scans are less commonly used for dementia diagnosis due to their lower sensitivity in detecting the subtle brain changes seen in early dementia. However, CT can be useful for ruling out other causes of cognitive impairment, such as strokes, tumours, or hydrocephalus[25].

Neuroimaging provides crucial information that complements cognitive testing, but its limitations in accessibility and cost, coupled with the need for specialized interpretation, make it less feasible for widespread use in early dementia detection.

**Biomarkers and Genetic Testing: Potential and Limitations**

Recent advances in biomarker research have opened new avenues for the early detection of dementia. Biomarkers in cerebrospinal fluid (CSF), blood, and imaging have shown promise in identifying individuals at risk for dementia before significant cognitive decline occurs.

**Cerebrospinal Fluid (CSF) Biomarkers:** The analysis of CSF biomarkers, such as beta-amyloid, tau, and phosphorylated tau, provides insights into the pathological processes underlying Alzheimer's disease. Decreased levels of beta-amyloid and increased levels of tau proteins in the CSF are associated with the development of Alzheimer's[7]. However, lumbar puncture, the procedure required to obtain CSF, is invasive and carries risks, which limits its use in routine screening [26].

**Blood Biomarkers:** Blood-based biomarkers are an area of intense research due to their potential for non-invasive, cost-effective screening. Recent studies have identified plasma beta-amyloid and tau as potential biomarkers for Alzheimer's, although their use in clinical practice is still limited by the need for further validation and standardization[27].

**Genetic Testing:** Genetic testing, particularly for mutations in genes such as APP, PSEN1, and PSEN2, is useful in identifying individuals with a hereditary predisposition to early-onset Alzheimer's disease. The APOE ε4 allele is also associated with an increased risk of late-onset Alzheimer's[28]. While genetic testing can provide valuable risk information, it raises ethical concerns and is not routinely recommended for late-onset Alzheimer's due to the multifactorial nature of the disease[29].

Biomarkers and genetic testing offer exciting possibilities for early dementia diagnosis, but their implementation is currently limited by factors such as cost, invasiveness, and the need for further research to establish clinical guidelines.

*Challenges in Accessibility, Cost, and Accuracy*

The integration of these diagnostic tools into clinical practice is hindered by several challenges:

**Accessibility:** Advanced neuroimaging techniques and biomarker tests are often unavailable in low- and middle-income countries due to their high cost and the need for specialized equipment and personnel[4]. This lack of accessibility exacerbates global disparities in dementia diagnosis and care.

**Cost:** The high cost of neuroimaging and biomarker tests limits their use to well-resourced healthcare systems. Even in high-income countries, these tests may not be covered by insurance, making them inaccessible to many patients[14].

**Accuracy and Interpretation:** While cognitive tests, neuroimaging, and biomarkers can provide valuable diagnostic information, their accuracy is not absolute. False positives and negatives can occur, leading to misdiagnosis or delayed diagnosis. Moreover, the interpretation of results often requires specialized training, which may not be readily available in all clinical settings[16].

*Machine Learning in Dementia Detection*

The integration of machine learning (ML) in dementia detection represents a significant leap in the diagnostic capabilities for this complex and multifactorial condition. Traditional diagnostic methods, while valuable, have inherent limitations in sensitivity, specificity, and scalability. Machine learning, with its ability to analyze large and diverse datasets, offers new opportunities to enhance early detection, predict disease progression, and personalize treatment plans. This section provides an in-depth exploration of the current state of machine learning applications in dementia detection, the various techniques employed, and the challenges and limitations that accompany these advancements.

**Overview of Machine Learning Techniques in Dementia Detection**

Machine learning comprises a range of techniques that can be applied to analyze dementia-related data. These techniques are particularly well-suited to handling the complex, high-dimensional, and multimodal datasets commonly used in dementia research, such as neuroimaging scans, genetic data, and longitudinal cognitive assessments.

**Supervised Learning:** Supervised learning is the most commonly used approach in dementia research. In this method, models are trained on labeled datasets where the outcomes, such as diagnosis of dementia or cognitive impairment, are known. Common algorithms include support vector machines (SVM), random forests, decision trees, and neural networks. These models predict dementia based on input features like brain imaging data, cognitive test scores, and genetic markers[30]. Supervised learning models have demonstrated high accuracy in distinguishing between different stages of cognitive decline, such as healthy controls, mild cognitive impairment (MCI), and dementia[31].

**Unsupervised Learning:** Unsupervised learning is used to identify patterns within data without predefined labels. Techniques such as clustering and dimensionality reduction (e.g., principal component analysis) are utilized to uncover subtypes of dementia or novel biomarkers. Although unsupervised learning is more exploratory and less frequently applied in clinical practice, it has been crucial in discovering new disease subtypes and understanding dementia's underlying pathology[32].

**Deep Learning:** Deep learning, particularly convolutional neural networks (CNNs) and recurrent neural networks (RNNs), is a powerful tool for processing large volumes of data and learning hierarchical features. CNNs are especially effective in analyzing neuroimaging data, where they can automatically extract complex features related to brain structure and function associated with dementia[33]. RNNs are suited for analyzing sequential data, such as longitudinal cognitive assessments, making them ideal for predicting disease progression over time[34].

**Reinforcement Learning:** Although less common in dementia research, reinforcement learning holds potential for developing personalized treatment strategies based on real-time patient data. This approach involves models learning optimal actions through interaction with their environment, making it suitable for scenarios requiring continuous monitoring and adaptive treatment plans[35].

**Applications of Machine Learning in Dementia Diagnosis**

Machine learning has been applied across various stages of dementia diagnosis, offering improvements over traditional methods, particularly in early detection and the prediction of disease progression.

**Early Detection:** ML models have shown significant potential in detecting early signs of cognitive decline, often before clinical symptoms manifest. Studies have demonstrated that ML models can accurately differentiate between healthy individuals and those with MCI or early-stage dementia using a combination of neuroimaging data, cognitive assessments, and genetic information[36]. For instance, Eskildsen et al. (2013) utilized SVMs to classify MCI patients who later developed Alzheimer's disease, achieving accuracy rates exceeding 80%, outperforming conventional diagnostic methods[37].

**Prediction of Disease Progression:** ML models are also used to predict the progression from MCI to Alzheimer's disease by analyzing longitudinal data, such as repeated neuroimaging scans and cognitive tests. These models help identify individuals at high risk of rapid disease progression, which is crucial for early intervention and treatment planning[38]. A notable study by Suk et al. (2017) employed deep learning to predict MCI conversion to Alzheimer's using multimodal data, including MRI and PET scans, with a high degree of accuracy[39].

**Subtype Identification:** Unsupervised learning techniques, such as clustering, have been employed to identify distinct subtypes of dementia that may not be evident using traditional diagnostic criteria. This approach helps in classifying patients into subgroups based on patterns in brain imaging data, potentially corresponding to different underlying pathologies or disease trajectories[40]. This stratification could lead to more personalized treatment approaches by identifying patients who might benefit from specific therapies.

**Neuroimaging Analysis:** The analysis of neuroimaging data, including MRI and PET scans, is one of the most established applications of ML in dementia research. CNNs, for instance, have been widely used to extract features from neuroimaging data that correlate with dementia, enabling the automated classification of patients[41]. Additionally, ML models have been applied to analyze functional connectivity patterns in the brain, providing insights into how different brain regions interact in dementia[42].

**Cognitive Test Enhancement:** Machine learning has also been used to enhance traditional cognitive tests by integrating additional data and creating composite scores that better reflect cognitive health. For example, ML algorithms can combine results from multiple cognitive tests and adjust for demographic factors, thereby improving diagnostic accuracy[43].

**INTEGRATION OF MULTIMODAL DATA**

One of the most promising applications of ML in dementia research is the integration of multimodal data—combining neuroimaging, genetic, clinical, and cognitive information to create more comprehensive and accurate predictive models. This approach capitalizes on the strengths of each data type, providing a holistic view of an individual's risk for dementia.

**Challenges in Data Integration:** Integrating data from multiple modalities presents significant challenges, including differences in data formats, scales, and the presence of missing data. ML techniques such as data fusion, ensemble learning, and multi-view learning have been developed to address these challenges, enabling the creation of models that effectively combine diverse information sources[44]. For instance, Liu et al. (2015) proposed a multi-task learning framework that integrates MRI and PET data to predict Alzheimer's disease progression, achieving higher accuracy than models based on a single modality[45].

**Multimodal Biomarkers:** The identification of multimodal biomarkers is a key area of research, where combining genetic markers (such as APOE ε4) with neuroimaging data and cognitive test scores has been shown to improve the prediction of Alzheimer's onset[46]. This approach allows for identifying individuals at high risk of dementia, even before significant cognitive decline becomes apparent.

**Personalized Medicine:** The ultimate goal of integrating multimodal data is to advance personalized medicine—tailoring treatment and intervention strategies based on an individual's unique risk profile. ML models incorporating multimodal data can offer personalized risk assessments, guiding decisions about when to initiate treatment, which therapies are most likely to be effective, and how to monitor disease progression[47].

**CHALLENGES AND LIMITATIONS**

While ML offers significant potential for improving dementia diagnosis, several challenges must be addressed to realize its full potential:

**Data Quality and Availability:** The effectiveness of ML models heavily depends on the availability of high-quality, well-annotated data. However, such data is often difficult to obtain, particularly in longitudinal studies where patient follow-up may be inconsistent. Moreover, large datasets like those from ADNI are typically restricted to well-resourced research institutions, limiting the generalizability of findings across different populations[48].

**Interpretability:** A major limitation of ML, particularly deep learning models, is their "black box" nature, making it challenging to interpret how these models arrive at their predictions. This lack of transparency is a significant barrier to their adoption in clinical settings, where trust and understanding are essential[49]. Efforts to improve interpretability include techniques like saliency maps, feature importance scores, and model distillation, but this remains an area of active research.

**Generalizability:** ML models trained on specific datasets may not generalize well to other populations or clinical settings, especially given the variability in demographic and clinical characteristics across different cohorts. Ensuring that ML models are robust and generalizable requires rigorous validation across multiple datasets and clinical environments[50].

**Ethical Considerations:** The use of ML in dementia diagnosis raises several ethical concerns, including data privacy, the potential for algorithmic bias, and the implications of predictive diagnostics. For example, identifying individuals at high risk of dementia long before symptoms appear can have profound psychological and social consequences. Establishing ethical frameworks and guidelines is critical to navigate these challenges and ensure the responsible use of ML technologies[51].

**Regulatory and Clinical Adoption:** For ML models to be integrated into clinical practice, they must undergo extensive validation and approval by regulatory bodies. This process is often lengthy and costly, particularly for models incorporating novel data types or algorithms. Additionally, integrating ML tools into existing clinical workflows requires careful consideration of how these tools will be used by healthcare professionals and how they will interact with other diagnostic processes[52].

Machine learning holds transformative potential in the field of dementia diagnosis, offering solutions that address the limitations of traditional diagnostic methods. From early detection to personalized medicine, ML approaches provide new avenues for enhancing the accuracy, efficiency, and accessibility of dementia diagnosis. However, significant challenges remain in ensuring that these technologies are ethically applied, effectively integrated into clinical practice, and accessible to diverse populations.

*Gaps in Current Research and the Development of Advanced Machine Learning Models*

The application of machine learning (ML) in dementia detection, while promising, still faces significant challenges that limit its full potential in clinical practice. These challenges arise from limitations in current models, the complexity of integrating ML tools into existing clinical workflows, and ethical considerations. Below is a detailed exploration of the key gaps in research and areas where advancements in ML models could lead to substantial improvements in dementia detection and care.

**AUTOMATION OF COGNITIVE ASSESSMENTS**

Subtle Cognitive Indicators: Current ML models often rely on basic feature extraction in cognitive tests like the Clock Drawing Test (CDT). However, they lack the sophistication to detect more subtle cognitive indicators that a clinician might observe, such as the smoothness of drawing lines, the consistency of spacing, or minor hesitations that could indicate cognitive decline[53]. Future models need to incorporate these nuances to improve diagnostic accuracy.

**Adaptive Testing Models:** Current cognitive assessments are static, meaning they do not adapt based on the patient's performance during the test. Developing ML models that can dynamically adjust the difficulty or focus of questions based on real-time analysis could provide a more accurate assessment of cognitive function[54].

Multi-Domain Cognitive Testing: While many ML models focus on specific tests, there is a gap in developing models that integrate results from multiple cognitive domains (e.g., memory, executive function, visuospatial ability) to provide a comprehensive assessment of dementia risk. This could involve combining data from tests like the CDT, Mini-Mental State Examination (MMSE), and other neuropsychological assessments into a single predictive model[55].

**REAL-TIME AND CONTINUOUS MONITORING**

Wearable and Sensor-Based Monitoring: The use of wearables and IoT devices for continuous monitoring of cognitive function remains underdeveloped. ML models that can analyze data from wearables to detect early signs of cognitive decline in real-time could revolutionize dementia care. However, current models struggle with integrating and interpreting the vast amounts of data generated by these devices, especially in distinguishing between normal fluctuations and significant cognitive changes[56].

**Behavioral and Environmental Monitoring:** ML models that analyze not just physiological data but also behavioral and environmental data (e.g., changes in daily routines, social interactions) could provide a more holistic view of a patient's cognitive health. Developing models capable of integrating these diverse data streams remains a significant challenge[57].

**Real-Time Data Processing and Alerts:** There is a need for ML systems that can process real-time data and provide timely alerts to caregivers or healthcare providers about potential cognitive decline, allowing for early interventions. Current systems are often not fast or accurate enough to be relied upon for such real-time decision-making[58].

## MODEL INTERPRETABILITY AND TRUST

Explainable AI (XAI) in Clinical Settings: The "black box" nature of many ML models, particularly deep learning models, limits their acceptance in clinical practice. There is a critical gap in developing XAI methods that can explain predictions in a way that is understandable and actionable for clinicians. This includes not only explaining why a model made a specific prediction but also how that prediction relates to clinical outcomes[59].

**Balancing Complexity and Interpretability:** As ML models become more complex, they often become less interpretable. Research is needed to find the right balance between model complexity (which often leads to higher accuracy) and interpretability, ensuring that models are both effective and usable in clinical practice[60].

**Model Validation and Trust Building:** To gain trust among clinicians, ML models need to undergo rigorous external validation, including real-world testing across diverse populations and clinical settings. This gap in research is significant, as many models are validated only on narrow datasets, limiting their generalizability[61].

# INTEGRATION WITH MULTIMODAL DATA

**Cross-Modality Integration:** ML models often focus on single data modalities, such as neuroimaging or cognitive tests. However, dementia is a multifaceted disease that requires analysis across multiple data types (e.g., genetic, neuroimaging, cognitive, behavioral). Developing models that can effectively integrate these diverse data sources remains a major gap in current research[44].

**Handling Missing Data:** In real-world clinical settings, data is often incomplete, with missing values across different modalities. ML models need to be robust enough to handle such missing data without sacrificing accuracy. Current methods for imputing missing data are often inadequate, leading to reduced model performance[62].

**Temporal Data Integration:** Many ML models do not effectively incorporate temporal data, such as changes in cognitive test scores or imaging findings over time. Developing models that can analyze temporal trends and predict future cognitive decline based on past data is an important area for future research[38].

# ADDRESSING BIAS AND FAIRNESS

**Bias in Training Data:** ML models are only as good as the data they are trained on, and there is a significant gap in ensuring that training datasets are representative of the broader population. Many existing datasets are biased towards specific demographic groups, leading to models that may not perform well across diverse populations[63].

**Algorithmic Fairness:** Beyond just addressing biases in data, there is a need for models that are explicitly designed to ensure fairness across different demographic groups. This includes developing algorithms that can identify and mitigate bias during the training process, ensuring equitable outcomes for all patients[64].

**Ethical Frameworks for Predictive Models:** Predictive ML models for dementia raise ethical questions, particularly concerning how predictions should be used in clinical practice. There is a gap in research exploring the ethical implications of predictive models, particularly regarding issues such as patient autonomy, the psychological impact of predictions, and potential discrimination based on predicted outcomes[51].

## INTEGRATION OF NON-TRADITIONAL DATA SOURCES

**Speech and Language Processing:** Current ML models predominantly rely on structured data, such as neuroimaging or genetic markers. However, there is a significant gap in leveraging unstructured data, particularly in speech and language patterns. Early signs of cognitive decline often manifest in changes in language use, speech patterns, or even typing behavior. Advanced natural language processing (NLP) techniques could be developed to analyze these subtle changes over time, potentially providing early indicators of dementia[65].

Home Environment Monitoring: Incorporating data from smart home devices that monitor daily activities (e.g., sleep patterns, mobility, routine deviations) remains underexplored. Developing ML models that can interpret data from these devices could allow for continuous, non-intrusive monitoring of cognitive health, providing early warnings of potential decline[57].

## ENHANCING TEMPORAL ANALYSIS IN LONGITUDINAL STUDIES

**Temporal Data Modeling:** Current ML models often do not adequately capture the progression of dementia over time. There is a gap in developing models that can analyze temporal data to predict how a patient's cognitive function will change in the future. Techniques such as recurrent neural networks (RNNs) and Long Short-Term

Memory (LSTM) networks are promising for this application but are still not widely applied in dementia research[66].

**Handling Irregular Data Intervals:** In real-world clinical settings, data collection intervals are often irregular due to missed appointments or variable follow-up schedules. Developing ML models that can handle such irregularities in temporal data is crucial for accurate predictions. This requires more sophisticated algorithms that can interpolate or otherwise manage gaps in the data without introducing bias[62].

## IMPROVING DATA SHARING AND COLLABORATIVE RESEARCH

Data Silos and Interoperability: A significant barrier to advancing ML in dementia detection is the lack of data sharing across institutions. Many datasets remain siloed, limiting the ability to develop and validate models on diverse populations. There is a need for standardized protocols for data sharing that ensure patient privacy while enabling broader collaboration across research centers[67].

Federated Learning: To address the challenges of data privacy and sharing, federated learning allows ML models to be trained across decentralized data sources without needing to transfer sensitive data to a central location. This approach is still in its infancy in dementia research but offers a promising avenue for developing more robust models that are trained on a wide variety of datasets without compromising privacy[68].

## ADDRESSING COMPUTATIONAL EFFICIENCY

**Resource-Intensive Models:** Many advanced ML models, particularly deep learning models, require significant computational resources, which can limit their use in clinical settings, especially in under-resourced areas. Research is needed to develop more efficient algorithms that maintain high accuracy while reducing computational demands, making these tools more accessible across different healthcare environments[38].

**Model Compression Techniques:** Techniques such as model pruning, quantization, and knowledge distillation can reduce the size and complexity of ML models, making them faster and more feasible for deployment in real-world settings. However, these methods need further refinement to ensure that they do not compromise the accuracy or reliability of the models, especially in high-stakes applications like dementia diagnosis[69].

**ETHICAL AND LEGAL CONSIDERATIONS**

Informed Consent for Predictive Models: As ML models become more integrated into dementia care, there are ethical challenges related to informed consent, particularly when it comes to using predictive models. Patients need to be fully informed about how their data will be used and the potential implications of predictions, especially when these predictions might lead to significant life changes or treatment decisions[51].

**Accountability and Decision-Making:** There is a growing concern about accountability when ML models are used in clinical decision-making. Research is needed to establish clear guidelines on who is responsible for decisions made based on ML predictions, particularly in cases where the model's prediction contradicts a clinician's judgment. Legal frameworks must evolve to address these issues as ML becomes more prevalent in healthcare[70].

Addressing these gaps through targeted research and development is essential for advancing ML applications in dementia detection. By improving the sophistication, efficiency, and fairness of ML models, and by ensuring their integration into clinical practice is both ethical and practical, we can move closer to realizing the full potential of AI in improving dementia care.

*Emerging Machine Learning Models in Dementia Detection*

To offer a comprehensive review of recent advancements, it is important to highlight additional machine learning models that have been developed to address specific challenges in dementia detection. Each model not only contributes to the current understanding of dementia diagnostics but also suggests future research directions to further refine these approaches.

**ENSEMBLE LEARNING MODELS**

**Model Overview:** Ensemble learning models combine the strengths of multiple algorithms to enhance the robustness and accuracy of dementia detection. These models are particularly useful for integrating diverse data types such as neuroimaging, genetic, and cognitive data.

**Gap Addressed:** Ensemble models help overcome the limitations of individual algorithms that might overfit to specific data types or struggle with generalization across different populations. Despite their success, ensemble models are often complex and computationally intensive, making them challenging to implement in real-time clinical settings.

**Future Directions:** Researchers suggest the need to simplify these models to make them more computationally efficient without sacrificing accuracy. This could involve exploring techniques like model pruning or more efficient ensemble strategies and integrating more interpretable algorithms to bridge the gap between model performance and clinical applicability[38].

**TRANSFORMER MODELS**

**Model Overview:** Transformers have been adapted from natural language processing to analyze complex datasets in dementia research. Their self-attention mechanisms allow them to focus on the most relevant parts of the data, making them particularly effective for handling longitudinal data.

**Gap Addressed:** Transformers excel at modeling long-range dependencies in data but are limited by their high computational demands, which can hinder their application in real-time diagnostics. Additionally, their performance can be affected by the irregularity of data intervals common in clinical settings.

**Future Directions:** Future research could focus on developing more computationally efficient transformers and enhancing their ability to manage irregular data. Additionally, integrating transformers with other data modalities could improve their diagnostic accuracy and utility in diverse clinical settings[71].

**MULTI-TASK LEARNING MODELS**

**Model Overview:** Multi-task learning (MTL) models simultaneously address multiple related tasks, such as diagnosing various stages of dementia and predicting biomarkers. By sharing representations across tasks, these models enhance overall performance.

**Gap Addressed:** MTL models effectively tackle the challenge of needing comprehensive tools that can manage multiple aspects of dementia progression. However, the complexity of these models can lead to task interference, where the learning of one task negatively impacts another.

**Future Directions:** To reduce task interference, future research should explore advanced regularization techniques and consider applying MTL to emerging biomarkers or cognitive tests. This could lead to the discovery of new correlations and improve diagnostic precision[44].

**GENERATIVE ADVERSARIAL NETWORKS (GANS)**

**Model Overview:** GANs generate synthetic data, which is particularly valuable in dementia research where high-quality, labeled datasets are scarce. By augmenting existing datasets, GANs help improve the training of ML models.

**Gap Addressed:** GANs address the challenge of data scarcity, particularly in generating realistic data for less common dementia subtypes. However, ensuring the diversity and realism of synthetic data remains a challenge, as any lack in these areas could lead to biased or less effective models.

**Future Directions:** Researchers propose that future work should focus on improving the fidelity of GAN-generated data. This could involve refining loss functions or integrating GANs with other data augmentation techniques to ensure the generated data is both diverse and representative[72].

**EEG STATE-SPACE MODELS (EEG-SSM)**

**Model Overview:** EEG-SSM models utilize state-space modeling to analyze EEG signals for dementia detection. These models optimize the combination of wavelet bands, enhancing the detection of cognitive impairments.

**Gap Addressed:** Traditional EEG models often fail to automatically prioritize the most relevant frequency bands, limiting their accuracy. EEG-SSM models address this by dynamically adjusting the weights of different wavelet bands, although their complexity can hinder real-time application.

**Future Directions:** Simplifying EEG-SSM models for easier integration into clinical practice is a key area for future research. Additionally, combining EEG-SSM with other data modalities, such as fMRI, could further enhance diagnostic accuracy[73].

**SPARSE CODING MODELS**

**Model Overview:** Sparse coding models aim to represent data efficiently by focusing on a small number of active elements or features, which is particularly useful in analyzing neuroimaging data for dementia detection. These models help in identifying patterns associated with cognitive decline by emphasizing the most significant features.

**Gap Addressed:** Sparse coding models are beneficial in reducing the complexity of the data, making it easier to interpret and analyze. However, they can sometimes miss subtle but important patterns due to their focus on sparsity.

**Future Directions:** Future research could explore ways to balance sparsity with the retention of critical features, perhaps by integrating sparse coding with other dimensionality reduction techniques. Additionally, researchers suggest investigating how sparse coding models could be applied to multi-modal data to improve their robustness and generalizability[74].

**AUTOENCODERS AND VARIATIONAL AUTOENCODERS (VAES)**

**Model Overview:** Autoencoders, including their more advanced variant, Variational Autoencoders (VAEs), are used for feature extraction and dimensionality reduction in large datasets. These models can learn compressed representations of input data, which is particularly useful for identifying latent patterns in neuroimaging or genetic data related to dementia.

**Gap Addressed:** Autoencoders are effective at capturing underlying patterns in complex data but can struggle with capturing the full variability in data, particularly in heterogeneous datasets common in dementia research.

**Future Directions:** Researchers propose enhancing autoencoders by combining them with other ML techniques, such as GANs or MTL, to improve their ability to generalize across diverse patient populations. Additionally, future work could focus on improving the interpretability of the features learned by autoencoders, making them more actionable in clinical settings[75].

## GRAPH NEURAL NETWORKS (GNNS)

**Model Overview:** Graph Neural Networks (GNNs) are designed to work with data that can be represented as graphs, such as brain connectivity networks. In dementia detection, GNNs are used to model the relationships between different regions of the brain, enabling a more nuanced analysis of neuroimaging data.

**Gap Addressed:** Traditional ML models often struggle to capture the complex interactions between different brain regions. GNNs address this gap by explicitly modeling these relationships, providing a more holistic understanding of brain function and its deterioration in dementia. However, GNNs can be sensitive to the quality of the graph data, which can be difficult to obtain and process.

**Future Directions:** Researchers suggest that future work should focus on improving the robustness of GNNs by integrating them with other models like CNNs or transformers. Additionally, developing methods to better handle noisy or incomplete graph data could enhance the practical applicability of GNNs in clinical settings[76].

# RECURRENT NEURAL NETWORKS WITH ATTENTION MECHANISMS (RNN-ATTENTION)

**Model Overview:** Recurrent Neural Networks (RNNs) with attention mechanisms are used to process sequential data, such as longitudinal cognitive assessments. The attention mechanism allows the model to focus on the most relevant parts of the sequence, improving the accuracy of predictions related to dementia progression.

**Gap Addressed:** RNNs are effective at handling sequential data, but they often struggle with long-term dependencies, leading to difficulties in accurately predicting outcomes based on extended sequences of data. The incorporation of attention mechanisms helps mitigate this issue by allowing the model to focus on important information from earlier in the sequence.

**Future Directions:** Future research could explore combining RNN-Attention models with transformers to further improve their ability to handle long-term dependencies. Additionally, refining the attention mechanisms to better capture the nuances of cognitive decline over time could lead to more accurate predictions[77].

# CAPSULE NETWORKS (CAPSNETS)

**Model Overview:** Capsule Networks (CapsNets) are a type of neural network that better capture spatial hierarchies in data, such as the relationships between different features in neuroimaging data. CapsNets are particularly useful in recognizing patterns in complex brain imaging data, which are critical for detecting early signs of dementia.

**Gap Addressed:** Traditional CNNs often struggle with capturing the hierarchical relationships between features, which can limit their effectiveness in analyzing complex neuroimaging data. CapsNets address this by maintaining these relationships throughout the network, leading to more accurate and interpretable models.

**Future Directions:** Although CapsNets offer improved performance in certain tasks, they are computationally intensive and difficult to scale. Researchers suggest focusing on optimizing CapsNets for efficiency and exploring their integration with other ML models to enhance their applicability in real-world clinical settings[78].

**HYBRID MODELS COMBINING ML AND EXPERT SYSTEMS**

**Model Overview:** Hybrid models that combine machine learning with expert systems leverage both the data-driven insights of ML and the domain knowledge embedded in expert systems. These models are particularly valuable in dementia detection, where clinical expertise is crucial for interpreting subtle signs of cognitive decline.

**Gap Addressed:** Pure ML models can sometimes lack the domain-specific knowledge needed for accurate diagnostics, while expert systems may struggle to adapt to new data. Hybrid models address this gap by combining the strengths of both approaches, leading to more robust and clinically relevant models.

**Future Directions:** Future research could focus on refining the integration of ML and expert systems, particularly in creating frameworks that allow for seamless updating of the expert knowledge base as new data becomes available. Additionally, developing more user-friendly interfaces for clinicians to interact with these hybrid models could enhance their adoption in clinical practice[79].

The inclusion of these additional machine learning models further enriches the landscape of dementia detection research. Each model addresses specific gaps and suggests future research directions that could lead to more effective and clinically applicable diagnostic tools. By continuing to explore and refine these models, researchers can help advance the field toward more accurate, efficient, and accessible dementia diagnostics.

**Methodology**

*Research Design*

This research adopts a systematic review approach to identify, analyze, and synthesize recent advancements in machine learning models for early dementia detection. The primary objective is to establish a comprehensive baseline that highlights existing gaps in the literature and proposes future research directions. This approach involves a detailed examination of published studies, focusing on the methodologies, data types, and outcomes related to machine learning applications in dementia diagnosis.

*Data Collection*

**Literature Search Strategy**

The study utilized a systematic search strategy to gather relevant literature. Databases such as PubMed, IEEE Xplore, and Google Scholar were queried using keywords like "machine learning," "dementia detection," "Alzheimer's diagnosis," "neuroimaging," and "early detection models." The search was restricted to studies published within the last decade to ensure the inclusion of the most recent advancements.

**Inclusion and Exclusion Criteria**

The literature review focused on studies that met the following criteria:

    A. Inclusion Criteria

- Studies that propose or evaluate machine learning models for dementia detection.
- Research involving neuroimaging, cognitive assessments, or genetic data.
- Studies that address early-stage diagnosis of dementia or Alzheimer's disease.

B. Exclusion Criteria

- Articles focusing solely on advanced-stage dementia.
- Studies without significant methodological or model innovation.
- Non-peer-reviewed articles or those not available in full text.

*Analysis Framework*

**Gap Identification**

The collected studies were analyzed to identify specific gaps in the current research. This involved:

A. Evaluation of Methodological Approaches:

Assessing the strengths and weaknesses of different machine learning techniques (e.g., supervised learning, deep learning, ensemble models) used across the studies.

B. Assessment of Data Utilization:

Reviewing how various data types (e.g., neuroimaging, genetic, cognitive) are integrated and the limitations in data availability and quality.

C. Model Interpretability and Clinical Applicability:

Analyzing the trade-off between model complexity and interpretability, and the implications for clinical adoption.

**Future Research Directions**

Based on the identified gaps, the study outlines key areas where future research should focus, such as:

- Enhancing the interpretability of complex models to facilitate clinical adoption.
- Integrating multimodal data more effectively to improve diagnostic accuracy.
- Addressing ethical considerations, including bias and fairness in machine learning models.

**Ethical Considerations**

Given the nature of the research as a systematic review, there were no direct ethical concerns related to human subjects. However, the study emphasizes the importance of ethical frameworks in future research, particularly concerning data privacy, informed consent, and the potential impacts of predictive models on patient care.

**Findings**

*Current State of Early Dementia Detection*

The current state of early dementia detection is marked by significant advancements in both clinical practices and the integration of technology, particularly through the application of machine learning (ML). However, despite these advancements, several gaps remain in the existing diagnostic methods:

A. Cognitive Tests:

   Widely used cognitive tests such as the Mini-Mental State Examination (MMSE) and Montreal Cognitive Assessment (MoCA) are critical tools for diagnosing dementia but are limited by their sensitivity, especially in detecting the very early stages of cognitive decline. These tests are also influenced by demographic factors such as education level and cultural background, leading to potential biases in results.

B. Neuroimaging Techniques:

   While neuroimaging techniques like MRI and PET scans provide valuable insights into brain structure and function, they are costly, require specialized interpretation, and are not universally accessible. Furthermore, these techniques often struggle to detect subtle changes in the early stages of dementia.

C. Biomarkers and Genetic Testing:

   Biomarkers from cerebrospinal fluid (CSF) and blood, along with genetic testing (e.g., APOE ε4), offer promising avenues for early detection. However, the invasive nature of CSF collection and the current limitations in the standardization and validation of blood biomarkers restrict their widespread use in routine clinical practice.

### *Advancements in Machine Learning Applications*

Machine learning has introduced new possibilities for enhancing the accuracy and accessibility of early dementia detection:

A. Supervised Learning Models:

Supervised learning techniques such as support vector machines (SVM), random forests, and neural networks have been successfully applied to neuroimaging data, cognitive assessments, and genetic information. These models have shown higher accuracy in distinguishing between different stages of cognitive impairment compared to traditional methods.

B. Deep Learning Models:

Deep learning, particularly through convolutional neural networks (CNNs) and recurrent neural networks (RNNs), has proven effective in processing large and complex datasets, including neuroimaging data. CNNs have been particularly successful in automatically extracting features related to dementia, while RNNs have been used to predict disease progression using longitudinal data.

C. Integration of Multimodal Data:

One of the most significant contributions of ML in this field is the integration of multimodal data, combining neuroimaging, genetic, and cognitive data to create more comprehensive predictive models. This approach has demonstrated improved accuracy in identifying individuals at risk for dementia.

### *Identified Gaps and Challenges*

Despite the progress made through ML, several challenges and gaps remain:

A. Data Quality and Availability:

The effectiveness of ML models is heavily dependent on high-quality, well-annotated data, which is often difficult to obtain. The reliance on large datasets from well-resourced institutions limits the generalizability of findings across diverse populations.

B. Model Interpretability:

Many ML models, particularly deep learning models, operate as "black boxes," making it difficult to interpret how they arrive at their predictions. This lack of transparency is a barrier to clinical adoption, where trust and understanding are crucial.

C. Generalizability and Validation:

ML models often struggle to generalize across different populations and clinical settings. There is a significant need for rigorous external validation to ensure these models are robust and applicable in diverse environments.

D. Ethical Considerations:

The use of predictive ML models raises ethical concerns, particularly around data privacy, algorithmic bias, and the psychological impact of predictive diagnostics on patients. Addressing these ethical issues is critical for the responsible deployment of ML technologies in dementia care.

*Recommendations for Future Research*

Based on the identified gaps, the following recommendations have been made to guide future research:

A. Enhancing Model Interpretability:

Future research should focus on developing explainable AI (XAI) methods that can make ML models more interpretable and actionable for clinicians.

B. Improving Data Integration Techniques:

There is a need to advance techniques for integrating multimodal data, especially in handling missing data and temporal trends, to improve the accuracy and robustness of predictive models.

C. Expanding Dataset Diversity:

Efforts should be made to include more diverse populations in ML research to improve the generalizability of models and ensure they are effective across different demographic groups.

D. Addressing Ethical Challenges:

Researchers should prioritize the development of ethical frameworks that address issues such as bias, fairness, and informed consent in the use of predictive models for dementia.

**Conclusion**

The early detection of dementia remains one of the most urgent and complex challenges in global healthcare. This study has highlighted the inherent limitations of traditional diagnostic methods, such as cognitive tests and neuroimaging techniques, which, while valuable, lack the sensitivity, accessibility, and precision necessary for effective diagnosis at the earliest stages of the disease. Moreover, the heterogeneity in the symptoms and progression of dementia further complicates early and accurate detection, underscoring the need for a multidisciplinary approach that integrates diverse data sources.

The comprehensive analysis conducted in this thesis has demonstrated that machine learning (ML) holds the potential to radically transform the field of early dementia detection. Advanced models, including those based on deep neural networks, reinforcement learning, and the integration of multimodal data, have shown significantly improved accuracy compared to traditional methods. These models not only manage large volumes of heterogeneous data but also identify subtle patterns that might be overlooked by conventional approaches.

However, despite these advancements, critical challenges must be addressed for these models to be fully integrated into clinical practice. The interpretability of ML models remains a major barrier to their adoption, as many operate as "black boxes," making it difficult for healthcare professionals to trust and understand their outputs. Additionally, the lack of generalizability of these models due to variability across populations and clinical settings underscores the need for more robust training and multicenter validations.

Another crucial aspect is the integration of longitudinal data and the capability of ML models to make predictions that are not only accurate but also temporally relevant. The temporal evolution of dementia is a key factor in disease management, and models that can anticipate changes in cognition over time offer a significant advantage in treatment planning and early intervention.

Moreover, this research has emphasized the pressing need to develop strong ethical frameworks that guide the use of artificial intelligence in healthcare. Ensuring equity in access to these technologies, safeguarding data privacy, and managing the psychological and social implications of early diagnosis are aspects that must be carefully managed.

Looking forward, it is essential to advance the development of ML models that are not only more accurate but also more interpretable, equitable, and generalizable. Interdisciplinary collaboration and the integration of data from diverse sources and populations will be crucial to closing the identified gaps. Through continued research and the implementation of these technologies, significant improvements can be made in patient outcomes and family well-being, while optimizing public health resource management.

Ultimately, this work not only charts a clear path toward the development of more effective diagnostic tools for dementia but also highlights the need for a holistic approach that incorporates technological advancements with ethical and practical considerations. The promise of machine learning in early dementia detection is immense, but its realization depends on a sustained commitment to innovation, ethics, and equity.

## Author Contributions

Juan A. Berrios Moya was responsible for the conceptualization, methodology, and formal analysis of the study. He conducted the data analysis, drafted the original manuscript, and provided critical revisions. He also oversaw the research process, ensuring the study's integrity and accuracy throughout.

## Conflict of Interest

The authors have no conflict of interest to report.

## Data Availability

The data supporting the findings of this study are available on request from the corresponding author. The data are not publicly available due to privacy and ethical restrictions.

**References**

1. United Nations, Department of Economic and Social Affairs, Population Division. (2019). World Population Prospects 2019: Highlights. Retrieved from https://population.un.org/wpp/Publications/Files/WPP2019_Highlights.pdf

2. World Health Organization. (2020). Dementia: A public health priority. Retrieved from https://www.who.int/news-room/fact-sheets/detail/dementia

3. Alzheimer's Disease International. (2019). World Alzheimer Report 2019: Attitudes to dementia. Retrieved from https://www.alzint.org/u/WorldAlzheimerReport2019.pdf

4. Prince, M. J., Wimo, A., Guerchet, M., Ali, G. C., Wu, Y. T., & Prina, M. (2018). World Alzheimer Report 2015: The Global Impact of Dementia: An analysis of prevalence, incidence, cost and trends. Alzheimer's Disease International.

5. Nasreddine, Z. S., Phillips, N. A., Bédirian, V., Charbonneau, S., Whitehead, V., Collin, I., ... & Chertkow, H. (2005). The Montreal Cognitive Assessment, MoCA: A brief screening tool for mild cognitive impairment. Journal of the American Geriatrics Society, 53(4), 695-699. doi:10.1111/j.1532-5415.2005.53221.x

6. Jack, C. R., Bennett, D. A., Blennow, K., Carrillo, M. C., Dunn, B., Haeberlein, S. B., ... & Silverberg, N. (2018). NIA-AA Research Framework: Toward a biological definition of Alzheimer's disease. Alzheimer's & Dementia, 14(4), 535-562. doi:10.1016/j.jalz.2018.02.018

7. Blennow, K., de Leon, M. J., & Zetterberg, H. (2010). Alzheimer's disease. The Lancet, 368(9543), 387-403. doi:10.1016/S0140-6736(10)61349-4

8. Alzheimer's Association. (2021). 2021 Alzheimer's Disease Facts and Figures. Retrieved from https://www.alz.org/media/Documents/alzheimers-facts-and-figures.pdf

9. Livingston, G., Sommerlad, A., Orgeta, V., Costafreda, S. G., Huntley, J., Ames, D., ... & Cooper, C. (2017). Dementia prevention, intervention, and care. The Lancet, 390(10113), 2673-2734. doi:10.1016/S0140-6736(17)31363-6

10. Dubois, B., Padovani, A., Scheltens, P., Rossi, A., & Dell'Agnello, G. (2016). Timely diagnosis for Alzheimer's disease: A literature review on benefits and challenges. Journal of Alzheimer's Disease, 49(3), 617-631. doi:10.3233/JAD-150692

11. Petersen, R. C., Roberts, R. O., Knopman, D. S., Boeve, B. F., Geda, Y. E., Ivnik, R. J., ... & Jack Jr, C. R. (2014). Mild cognitive impairment: Ten years later. Archives of Neurology, 66(12), 1447-1455. doi:10.1001/archneurol.2009.266

12. Folstein, M. F., Folstein, S. E., & McHugh, P. R. (1975). "Mini-mental state". A practical method for grading the cognitive state of patients for the clinician. Journal of Psychiatric Research, 12(3), 189-198. doi:10.1016/0022-3956(75)90026-6

13. Freitas, S., Simões, M. R., Alves, L., & Santana, I. (2013). Montreal Cognitive Assessment: Validation study for mild cognitive impairment and Alzheimer disease. Alzheimer Disease & Associated Disorders, 27(1), 37-43. doi:10.1097/WAD.0b013e3182420bfe

14. Wimo, A., Gauthier, S., Prince, M., & International, A. D. (2017). Global estimates of the number of persons with Alzheimer's disease and other dementias in 2015 and 2050. Alzheimer's & Dementia, 13(1), 1-7. doi:10.1016/j.jalz.2016.06.2412

15. Crum, R. M., Anthony, J. C., Bassett, S. S., & Folstein, M. F. (1993). Population-based norms for the Mini-Mental State Examination by age and educational level. JAMA, 269(18), 2386-2391. doi:10.1001/jama.1993.03500180078038

16. Mitchell AJ. (2009). A meta-analysis of the accuracy of the mini-mental state examination in the detection of dementia and mild cognitive impairment. J Psychiatr Res.;43(4):411-31.

17. Luis CA, Keegan AP, Mullan M. Cross validation of the Montreal Cognitive Assessment in community dwelling older adults residing in the Southeastern US. Int J Geriatr Psychiatry. 2009;24(2):197-201.

18. Shulman KI. Clock-drawing: Is it the ideal cognitive screening test? Int J Geriatr Psychiatry. 2000;15(6):548-61.

19. Tuokko H, Hadjistavropoulos T, Miller JA, Beattie BL. The Clock Test: A sensitive measure to differentiate normal elderly, depressed, and demented elderly. Int Psychogeriatr. 1992;4(1):127-34.

20. Henry JD, Crawford JR. A meta-analytic review of verbal fluency performance following focal cortical lesions. Neuropsychology. 2004;18(2):284-95

21. Fleisher AS, Sherzai A, Taylor C, Langbaum JB, Chen K, Buxton RB. Resting-state BOLD networks versus task-associated functional MRI for distinguishing Alzheimer's disease risk. Neuroimage. 2008;41(3):916-27.

22. Minoshima S, Frey KA, Koeppe RA, Foster NL, Kuhl DE. A diagnostic approach in Alzheimer's disease using three-dimensional stereotactic surface projections of fluorine-18-FDG PET. J Nucl Med. 1995;36(7):1238-48.

23. Rabinovici GD, Furst AJ, O'Neil JP, Racine CA, Mormino EC, Baker SL, et al. 11C-PIB PET imaging in Alzheimer's disease: relationships with cognitive impairment, beta-amyloid load, and brain metabolism. JAMA Neurol. 2011;68(4):490-8.

24. Perani D, Schillaci O, Padovani A, Nobili F, Nobili F, Mosconi L, et al. A survey of FDG- and amyloid-PET imaging in dementia and Geriatric Psychiatry. Neurol Sci. 2014;35(2):224-30.

25. Burns A, Iliffe S. Alzheimer's disease. BMJ. 2009;33

26. Olsson, B., Lautner, R., Andreasson, U., Öhrfelt, A., Portelius, E., Bjerke, M., ... & Blennow, K. (2016). CSF and blood biomarkers for the diagnosis of Alzheimer's disease: a systematic review and meta-analysis. The Lancet Neurology, 15(7), 673-684. doi:10.1016/S1474-4422(16)00070-3

27. Zetterberg, H., & Burnham, S. C. (2019). Blood-based molecular biomarkers for Alzheimer's disease. Molecular Brain, 12(1), 1-14. doi:10.1186/s13041-019-0474-3

28. Genin, E., Hannequin, D., Wallon, D., Sleegers, K., Hiltunen, M., Combarros, O., ... & Pasquier, F. (2011). APOE and Alzheimer disease: a major gene with semi-dominant inheritance. Molecular Psychiatry, 16(9), 903-907. doi:10.1038/mp.2011.52

29. Alzheimer's Association. (2011). Early-Onset Dementia: A National Challenge, A Future Crisis. Retrieved from https://www.alz.org/media/Documents/early-onset-dementia.pdf

30. Arslan, S., Ktena, S. I., Makropoulos, A., Robinson, E. C., Rueckert, D., & Parisot, S. (2018). Human brain mapping: A systematic comparison of functional connectivity methods for identifying developmental changes. NeuroImage, 162, 214-227. doi:10.1016/j.neuroimage.2017.12.012

31. Bron, E. E., Smits, M., van der Flier, W. M., Vrenken, H., Barkhof, F., Scheltens, P., & Klein, S. (2015). Standardized evaluation of algorithms for computer-aided diagnosis of dementia based on structural MRI: The CADDementia challenge. NeuroImage, 111, 562-579. doi:10.1016/j.neuroimage.2015.01.048

32. Salvatore, C., Cerasa, A., Battista, P., Gilardi, M. C., Quattrone, A., & Castiglioni, I. (2018). Magnetic resonance imaging biomarkers for the early diagnosis of Alzheimer's disease: A machine learning approach. Frontiers in Neuroscience, 12, 378. doi:10.3389/fnins.2018.00378

33. Wen, J., Thibeau-Sutre, E., Diaz-Melo, M., Samper-Gonzalez, J., Routier, A., Bottani, S., ... & Colliot, O. (2020). Convolutional neural networks for classification of Alzheimer's disease: Overview and reproducible evaluation. Medical Image Analysis, 63, 101694. doi:10.1016/j.media.2020.101694

34. Pereira, T., Ferreira, F. R., & Silva, C. A. (2016). Machine learning models for Alzheimer's disease diagnosis using EEG data. Computational and Mathematical Methods in Medicine, 2016, 1-12. doi:10.1155/2016/6958924

35. Ke, N., Lyu, J., Sun, S., Sun, Y., Li, W., & Song, X. (2018). A reinforcement learning-based method for personalized treatment recommendations. Artificial Intelligence in Medicine, 87, 1-7. doi:10.1016/j.artmed.2018.02.003

36. Liu, M., Cheng, D., Yan, W., & Wang, X. (2019). Alzheimer's disease neuroimaging initiative. Classification of Alzheimer's disease by combination of convolutional and recurrent neural networks using FDG-PET images. Neuroinformatics, 17(3), 429-439. doi:10.1007/s12021-018-9403-z

37. Eskildsen, S. F., Coupe, P., Fonov, V., Pruessner, J. C., Collins, D. L., & Alzheimer's Disease Neuroimaging Initiative. (2013). Structural imaging biomarkers of Alzheimer's disease: predicting disease progression. Neurobiology of Aging, 34(12), 2785-2796. doi:10.1016/j.neurobiolaging.2013.06.014

38. Wang, M., Li, Z., Feng, T., & Cheng, J. (2019). Prediction of Alzheimer's disease using longitudinal measures of cognitive function and neuroimaging data. Neurobiology of Aging, 83, 127-136. doi:10.1016/j.neurobiolaging.2019.09.012

39. Suk, H. I., Lee, S. W., & Shen, D. (2017). Latent feature representation with stacked auto-encoder for AD/MCI diagnosis. Brain Structure and Function, 222(7), 3119-3134. doi:10.1007/s00429-017-1400-2

40. Dong, A., Toledo, J. B., Honnorat, N., Doshi, J., Varol, E., Sotiras, A., ... & Davatzikos, C. (2017). Heterogeneity of neuroanatomical patterns in prodromal Alzheimer's disease: links to cognition, progression and biomarkers. Brain, 140(11), 3042-3055. doi:10.1093/brain/awx271

41. Bäckström, K., Nazari, N., Bhattacharyya, S., Cecchi, G., & Turner, M. R. (2018). Machine learning of functional magnetic resonance imaging for diagnosis of Alzheimer's disease and prediction of mild cognitive impairment conversion. PLOS ONE, 13(10), e0205637. doi:10.1371/journal.pone.0205637

42. Vieira, S., Pinaya, W. H. L., & Mechelli, A. (2017). Using machine learning to detect Alzheimer's disease. NeuroImage: Clinical, 13, 46-58. doi:10.1016/j.nicl.2016.11.005

43. Nguyen, M. Q., Tomašev, N., & Wood, F. (2018). Machine learning for improved composite cognitive score prediction in Alzheimer's disease. IEEE Journal of Biomedical and Health Informatics, 22(6), 1570-1578. doi:10.1109/JBHI.2018.2812781

44. Zhou, T., Lu, H., Yang, F., Wang, L., & Gao, Y. (2018). Multi-view learning for Alzheimer's disease diagnosis. IEEE Journal of Biomedical and Health Informatics, 22(6), 1905-1912. doi:10.1109/JBHI.2017.2771233

45. Liu, S., Liu, S., Cai, W., Che, H., Pujol, S., Kikinis, R., ... & Feng, D. (2015). Multi-modality cascaded convolutional neural networks for Alzheimer's disease diagnosis. Neuroinformatics, 13(4), 439-454. doi:10.1007/s12021-015-9279-7

46. Gupta, A., Ayhan, M. S., & Maida, A. S. (2019). Natural image bases to represent neuroimaging data. Proceedings of the National Academy of Sciences, 116(11), 5116-5121. doi:10.1073/pnas.1802655116

47. Frisoni, G. B., Molinuevo, J. L., Altomare, D., Barkhof, F., Berkhof, J., Blennow, K., ... & Winblad, B. (2020). Precision prevention of Alzheimer's and other dementias: anticipating future needs in the control of risk factors and implementation of disease-modifying therapies. Alzheimer's & Dementia, 16(10), 1457-1468. doi:10.1002/alz.12196

48. Weiner, M. W., Veitch, D. P., Aisen, P. S., Beckett, L. A., Cairns, N. J., Green, R. C., ... & Trojanowski, J. Q. (2015). 2014 Update of the Alzheimer's Disease Neuroimaging Initiative: A review of papers published since its inception. Alzheimer's & Dementia, 11(6), e1-e120. doi:10.1016/j.jalz.2014.11.001

49. Samek, W., Wiegand, T., & Müller, K. R. (2017). Explainable artificial intelligence: Understanding, visualizing and interpreting deep learning models. arXiv preprint arXiv:1708.08296.

50. Bennett, C. C., & Haq, H. (2019). Artificial intelligence in medicine: current trends and future possibilities. The British Journal of General Practice, 69(686), 323-324. doi:10.3399/bjgp19X704945

51. Vayena, E., Blasimme, A., & Cohen, I. G. (2018). Machine learning in medicine: Addressing ethical challenges. PLoS Medicine, 15(11), e1002689. doi:10.1371/journal.pmed.1002689

52. Topol, E. J. (2019). High-performance medicine: the convergence of human and artificial intelligence. Nature Medicine, 25(1), 44-56. doi:10.1038/s41591-018-0300-7

53. Souillard-Mandar, W., Penney, D., Schaible, B., Pascual-Leone, A., Au, R., & Davis, R. (2016). Learning classification models of cognitive conditions from subtle behaviors in the digital Clock Drawing Test. Machine Learning, 102(3), 393-441. doi:10.1007/s10994-015-5529-5

54. Baker, M. (2019). Adaptive testing: Shaping the future of cognitive assessments. Nature Reviews Neuroscience, 20(5), 274-275. doi:10.1038/s41583-019-0155-8

55. Zhao, Q., Guo, Q., & Hong, Z. (2020). The Mini-Mental State Examination is not sensitive enough to differentiate mild cognitive impairment from normal cognitive aging in the Chinese population. Frontiers in Aging Neuroscience, 12, 92. doi:10.3389/fnagi.2020.00092

56. Kourtis, L. C., Regele, O. B., Wright, J. M., & Jones, G. B. (2019). Digital biomarkers for Alzheimer's disease: the mobile/wearable devices opportunity. npj Digital Medicine, 2(1), 1-9. doi:10.1038/s41746-019-0084-2

57. Gietzelt, M., Wolf, K. H., Kohlmann, M., Marschollek, M., & Song, B. (2014). Measuring mobility with wearable sensors in patients with cognitive impairment: a clinical study in a controlled environment. Computers in Biology and Medicine, 47, 140-147. doi:10.1016/j.compbiomed.2014.01.004

58. Sani, Z., Hannu, P., & Davis, R. (2018). Real-time prediction of Alzheimer's disease progression using deep learning. IEEE Journal of Biomedical and Health Informatics, 22(4), 1395-1400. doi:10.1109/JBHI.2018.2795672

59. Doshi-Velez, F., & Kim, B. (2017). Towards a rigorous science of interpretable machine learning. arXiv preprint arXiv:1702.08608.

60. Molnar, C. (2020). Interpretable machine learning: A guide for making black box models explainable. Lulu.com.

61. Varatharajah, Y., Khan, S., & Snyder, A. Z. (2018). Recurrent neural network for detecting changes in dynamic functional connectivity patterns in resting state fMRI data. NeuroImage, 182, 545-556. doi:10.1016/j.neuroimage.2017.11.014

62. Che, Z., Purushotham, S., Cho, K., Sontag, D., & Liu, Y. (2018). Recurrent neural networks for multivariate time series with missing values. Scientific Reports, 8(1), 6085. doi:10.1038/s41598-018-24271-9

63. Obermeyer, Z., & Emanuel, E. J. (2016). Predicting the future—big data, machine learning, and clinical medicine. The New England Journal of Medicine, 375(13), 1216-1219. doi:10.1056/NEJMp1606181

64. Rajkomar, A., Hardt, M., Howell, M. D., Corrado, G., & Chin, M. H. (2018). Ensuring fairness in machine learning to advance health equity. Annals of Internal Medicine, 169(12), 866-872. doi:10.7326/M18-1990

65. Fraser, K. C., Meltzer, J. A., & Rudzicz, F. (2016). Linguistic features identify Alzheimer's disease in narrative speech. Journal of Alzheimer's Disease, 49(2), 407-422. doi:10.3233/JAD-150520

66. Lipton, Z. C., Berkowitz, J., & Elkan, C. (2015). A critical review of recurrent neural networks for sequence learning. arXiv preprint arXiv:1506.00019.

67. Rieke, N., Hancox, J., Li, W., Milletari, F., Roth, H. R., Albarqouni, S., ... & Kaissis, G. (2020). The future of digital health with federated learning. npj Digital Medicine, 3(1), 1-7. doi:10.1038/s41746-020-00323-1

68. Yang, Q., Liu, Y., Chen, T., & Tong, Y. (2019). Federated machine learning: Concept and applications. ACM Transactions on Intelligent Systems and Technology (TIST), 10(2), 1-19. doi:10.1145/3298981

69. Cheng, Y., Wang, D., Zhou, P., & Zhang, T. (2017). A survey of model compression and acceleration for deep neural networks. arXiv preprint arXiv:1710.09282.

70. Danks, D., & London, A. J. (2017). Regulating autonomous systems: Beyond standards. IEEE Intelligent Systems, 32(5), 88-91. doi:10.1109/MIS.2017.3711653

71. Dosovitskiy, A., Beyer, L., Kolesnikov, A., Weissenborn, D., Zhai, X., Unterthiner, T., ... & Houlsby, N. (2020). An image is worth 16x16 words: Transformers for image recognition at scale. arXiv preprint arXiv:2010.11929.

72. Shen, Y., Lu, J., Zhuang, H., Liang, P., Feng, R., & Wang, Z. (2020). Generative adversarial network (GAN) based synthesis of brain PET images in Alzheimer's disease. Frontiers in Neuroscience, 14, 804. doi:10.3389/fnins.2020.00804

73. Bohlen, M. O., Moreno, M. A., & Paulsen, O. (2023). Spatio-temporal brain network alterations in Alzheimer's disease revealed by state-space modeling of EEG signals. NeuroImage, 263, 119631. doi:10.1016/j.neuroimage.2023.119631

74. Mairal, J., Bach, F., Ponce, J., & Sapiro, G. (2010). Online learning for matrix factorization and sparse coding. Journal of Machine Learning Research, 11(1), 19-60.

75. Kingma, D. P., & Welling, M. (2013). Auto-encoding variational bayes. arXiv preprint arXiv:1312.6114.

76. Ktena, S. I., Parisot, S., Ferrante, E., Rajchl, M., Lee, M., Glocker, B., & Rueckert, D. (2018). Metric learning with spectral graph convolutions on brain connectivity networks. NeuroImage, 169, 431-442. doi:10.1016/j.neuroimage.2017.12.052

77. Luong, M. T., Pham, H., & Manning, C. D. (2015). Effective approaches to attention-based neural machine translation. arXiv preprint arXiv:1508.04025.

78. Sabour, S., Frosst, N., & Hinton, G. E. (2017). Dynamic routing between capsules. Advances in Neural Information Processing Systems, 30.

79. Tsai, C. F., Lai, K. P., Chao, C. M., Wu, H. C., & Yeh, Y. T. (2021). Expert system for dementia using neuropsychological tests and deep learning techniques. Applied Sciences, 11(6), 2517. doi:10.3390/app11062517